# Review of PID Controller Applications for UAVs

**Hans Oersted [1]\*, Yudong Ma [2]**

[1] Zhejiang University; 3130102046@zju.edu.cn
[2] The University of Tokyo; mayudong200333@gmail.com
\* Correspondence

**Abstract:** Unmanned Aerial Vehicles (UAVs) have gained widespread recognition for their diverse applications, ranging from surveillance to delivery services. Among the various control algorithms employed to stabilize and navigate UAVs, the Proportional-Integral-Derivative (PID) controller stands out as a classical yet robust solution. This review provides a comprehensive examination of PID controller applications in the context of UAVs, addressing their fundamental principles, dynamics modeling, stability control, navigation tasks, parameter tuning methods, challenges, and future directions.

## 1. Introduction

The evolution of Unmanned Aerial Vehicles (UAVs) represents a captivating journey through decades of technological advancements [1,2], gradually transforming these aerial platforms from rudimentary beginnings to highly sophisticated systems with diverse applications. Concurrently, the development of the Proportional-Integral-Derivative (PID) controller has played a pivotal role in shaping the control systems that govern UAVs, particularly in the context of quadrotor dynamics [3].

The roots of UAVs can be traced back to the early 20th century, marked by the development of radio-controlled aircraft during World War I. These early UAVs were primarily experimental and military-focused. The technological leaps following World War II ushered in a new era of UAVs, with the advent of reconnaissance drones. These UAVs were often used for surveillance purposes, marking the beginning of their integration into military operations [4].

The late 20th century witnessed significant strides in UAV technology, driven by advancements in computing, communication, and miniaturization. This period saw the emergence of more sophisticated UAVs with improved range, endurance, and payload capacities.

In the past few decades, UAVs transitioned from predominantly military use to a plethora of civilian applications. Aerial photography, agriculture, environmental monitoring, and search and rescue operations are just a few examples of how UAVs have become integral to various industries.

The concept of feedback control, the foundation of PID control, dates back to the 19th century. However, it wasn't until the mid-20th century that the PID controller as we know it today began to take shape [5]. PID controllers found widespread use in industrial automation, where their simplicity and effectiveness made them a preferred choice for regulating processes.

Ongoing advancements in control theory, coupled with the rise of computational capabilities, led to refinements in PID algorithms. These improvements addressed specific challenges and enhanced performance in various applications [6].

As quadrotors gained popularity in the early 21st century, PID controllers emerged as a natural choice for stabilizing these inherently unstable systems. Initial research focused on basic stability control. The last decade witnessed a shift towards leveraging PID controllers for navigation tasks in quadrotors. PID algorithms were adapted for trajectory control, path tracking, and response to external disturbances.

The current landscape involves addressing challenges such as nonlinearity and external factors. Additionally, researchers are exploring the integration of advanced techniques like adaptive control, deep learning, and reinforcement learning to enhance PID control in quadrotors.

**2. PID Controller**

The Proportional-Integral-Derivative (PID) controller serves as a cornerstone in control systems, providing a straightforward yet effective approach to regulating a system's behavior. At its core, a PID controller operates by adjusting the system output based on three components: Proportional (P) component responds to the current error, ensuring that the system output is directly proportional to the difference between the desired and actual values. Integral (I) component considers the accumulation of past errors, aiming to eliminate any steady-state error that may persist over time. Derivative (D) component anticipates future system behavior by examining the rate of change of the error, helping to dampen oscillations and improve stability.

In the context of quadrotor control, PID controllers are employed to stabilize the system, ensuring that it responds promptly to external disturbances and maintains a desired orientation and position during flight. The adaptability and simplicity of PID controllers make them well-suited for real-time applications, contributing to their widespread use in quadrotor dynamics.

Quadrotors, with their complex dynamics and nonlinearities, pose significant challenges for control system design. Feedback linearization emerges as a powerful technique to address these challenges by transforming the inherently nonlinear quadrotor dynamics into a linear and controllable form [7,8]. This section explores the principles of feedback linearization and its profound implications for enhancing the control capabilities of quadrotor systems [9,10].

Feedback linearization operates on the premise of transforming a nonlinear system into a linear one through appropriate feedback control [11,12]. The core idea is to introduce a control law that cancels out the nonlinearities, thereby rendering the system linear and amenable to well-established linear control techniques [13]. This process involves carefully selecting control inputs and state transformations to achieve a linearized representation of the system.

In the context of quadrotor control [14], feedback linearization proves invaluable due to the complex interactions between the multiple rotors, torques, and external disturbances [15,16]. The rotational and translational dynamics of a quadrotor are inherently nonlinear, making traditional control approaches challenging.

Feedback linearization is applied to quadrotors by carefully choosing the control inputs and manipulating the system's states. By canceling out the nonlinear terms in the dynamics equations, the quadrotor's behavior can be transformed into a linear form [17], allowing for the application of conventional linear control techniques.

The linearized quadrotor model obtained through feedback linearization enables the use of linear control strategies, such as Proportional-Integral-Derivative (PID) controllers [18,19], for precise and agile control. Beyond stabilization, this technique facilitates advanced control tasks, including trajectory tracking, path following, and intricate maneuvering.

One notable advantage is the simplification of control design [20], as linear controllers are generally easier to analyze and tune. This not only enhances the ease of implementation but also allows for the application of established control methodologies to achieve desired performance.

While feedback linearization offers significant benefits, challenges arise in practice [21]. Accurate modeling of the quadrotor dynamics is essential for successful linearization, and uncertainties or variations from the modeled dynamics can affect performance. Additionally, robustness to external disturbances and parameter uncertainties must be carefully considered in the control design.

Ongoing research in quadrotor control continues to explore advancements in feedback linearization techniques, aiming to enhance robustness [22,23], adaptability, and real-time implementation. The integration of machine learning and adaptive control with feedback linearization represents a promising avenue for further improving quadrotor control capabilities.

Feedback linearization stands as a transformative approach in the realm of quadrotor control, offering a means to tame the inherent nonlinearities and unlock enhanced control capabilities [24]. By providing a linearized representation, this technique facilitates the application of traditional control strategies, paving the way for more sophisticated and precise control of quadrotor systems in various applications.

### 3. Challenges in PID Controller on Stabilizing UAVs

Quadrotors exhibit intricate nonlinear dynamics, arising from the complex interactions between multiple rotors and their influence on both translational and rotational motion. PID controllers, designed for linear systems, may struggle to accurately model and address the nonlinear behaviors of quadrotors. The challenge lies in reconciling the inherent nonlinearities with the linear assumptions of PID control, potentially leading to suboptimal performance and the need for advanced control strategies.

The strong coupling between different axes of quadrotor motion complicates control design. PID controllers, typically developed under the assumption of single-input, single-output linearity, may face difficulties in effectively decoupling these motions. The interdependence between control loops can lead to undesired interactions, impacting the stability and precision of the quadrotor. Addressing axis coupling challenges requires careful consideration in control algorithm design and tuning.

Quadrotors operate in diverse and dynamic environments, subject to fluctuations in wind conditions, payload weights, and external disturbances. PID controllers, traditionally tuned for specific operating conditions, may lack the adaptability to ensure consistent and robust performance across varying scenarios. Achieving a balance between responsive control and adaptability to different operating conditions remains a significant challenge in PID-based control strategies for quadrotors.

The presence of actuators, such as motors and propellers, introduces limitations in terms of response time and saturation in quadrotor systems. Aggressive control signals from improperly tuned PID controllers can push actuators to their limits, resulting in saturation and diminishing the effectiveness of the control system. Integrating the dynamics of actuators into the control design becomes essential to ensure stable and reliable quadrotor control.

External disturbances, including sudden wind gusts, pose a substantial challenge to PID controllers in maintaining stability and trajectory control. PID controllers, primarily designed for nominal conditions, may struggle to adapt quickly and effectively to external forces. Robust strategies that can anticipate and counteract external disturbances, ensuring stability and trajectory tracking, remain an ongoing research focus for improving PID-based control of quadrotors.

Tuning PID parameters for quadrotors can be a complex and intricate task. The dynamic and nonlinear nature of quadrotor systems requires careful consideration of various factors during the tuning process. Achieving an optimal balance between stability and responsiveness necessitates expertise in control theory and an in-depth understanding of the quadrotor's dynamics, adding complexity to the implementation of PID controllers in practical applications [25].

While PID controllers excel at basic stabilization, their limitations become apparent when tackling advanced control tasks. Integrating advanced control methods, such as model predictive control, adaptive control, or machine learning, with PID controllers requires addressing issues of system integration, real-time implementation, and seamless cooperation between different control layers. Balancing the simplicity of PID with the sophistication of advanced control strategies poses a multifaceted challenge in the pursuit of comprehensive quadrotor control.

In conclusion, addressing these challenges in applying PID controllers to quadrotor systems is crucial for unlocking their full potential across diverse operating conditions and control tasks [26]. The ongoing research aims to devise innovative solutions that enhance the adaptability, robustness, and overall performance of PID-based control strategies for quadrotors.

## 4. Stability Considerations in Quadrotor PID Control

The stability of quadrotors using PID controllers is a critical aspect that influences their overall performance. PID controllers are designed to stabilize systems by adjusting control inputs based on the error between the desired and actual states [27,28]. In the context of quadrotors, PID controllers play a key role in stabilizing the inherent instabilities of these aerial systems, enabling them to maintain a desired orientation and position during flight.

The proportional, integral, and derivative components of PID controllers contribute to stabilizing quadrotors in the following ways: Proportional (P): The proportional term responds to the current error, providing immediate correction to stabilize the quadrotor. It is crucial for counteracting disturbances and maintaining a consistent response. Integral (I): The integral term addresses steady-state errors that may persist over time, preventing long-term drift in the quadrotor's position or orientation. Derivative (D): The derivative term anticipates future behavior by considering the rate of change of the error. It helps dampen oscillations and enhances the overall stability of the quadrotor [5].

Despite the fundamental role of PID controllers in stabilizing quadrotors, challenges exist that can compromise stability under certain conditions.

The nonlinear dynamics and strong coupling between different axes can introduce challenges in accurately modeling and addressing the system's behavior. In situations where nonlinearities are pronounced, PID controllers may struggle to provide adequate stabilization.

Quadrotors are often subjected to extreme operating conditions, such as high wind speeds or abrupt changes in payload weight [29]. PID controllers, tuned for specific conditions, may exhibit reduced stability or responsiveness in the face of these extremes.

Variability in quadrotor parameters, such as motor characteristics or sensor accuracy, can impact the effectiveness of PID controllers [30]. If these variations are not adequately accounted for in the controller design or tuning process, stability may be compromised.

Actuator limitations, such as saturation and dynamics, can affect stability, especially if PID controllers push the actuators beyond their operational limits. Managing these constraints becomes crucial for maintaining stability.

To address stability challenges in quadrotor PID control, researchers are exploring advanced control strategies and tuning methodologies. Techniques such as adaptive control, where the controller adapts to changing system dynamics, and robust control, which accounts for uncertainties and disturbances, offer promising avenues to enhance stability under challenging conditions [31,32].

Furthermore, the integration of machine learning approaches can aid in developing adaptive controllers that learn and adjust to evolving quadrotor dynamics, contributing to improved stability in a wider range of scenarios.

While PID controllers provide a robust foundation for stabilizing quadrotors, acknowledging and addressing the challenges to stability is crucial for unlocking their full potential across diverse operating conditions and disturbances. Ongoing research efforts continue to advance control methodologies, ensuring the stability and reliability of quadrotor systems in real-world applications.

## 5. Conclusions

This review has embarked on a comprehensive exploration of the historical evolution, control strategies, and challenges surrounding the integration of PID controllers in the dynamic realm of quadrotor systems. The historical journey unveiled the transformative growth of Unmanned Aerial Vehicles (UAVs) from their early military roots to their present-day versatile applications in both civilian and defense sectors.

PID controllers, with their simplicity and efficacy, have emerged as fundamental tools in stabilizing quadrotors, allowing these agile aerial platforms to perform tasks ranging from basic stabilization to advanced navigation. The fundamental principles of Proportional-Integral-Derivative control provide a solid foundation for addressing the inherent instabilities of quadrotors, enabling precise control in various environments and operating conditions.

However, the discussion also shed light on the multifaceted challenges associated with PID control in quadrotor systems. The nonlinearities and complexities inherent in quadrotor dynamics, coupled with issues of coupling between different motion axes, highlight the need for nuanced control strategies. Variability in operating conditions, saturation in actuator dynamics, and the ever-present influence of external disturbances pose additional hurdles in achieving robust and adaptive control.

The exploration extended to sophisticated techniques like feedback linearization, offering a means to transform the nonlinear dynamics of quadrotors into a more manageable linear form. This approach, combined with PID controllers, opens avenues for advanced control tasks, showcasing the constant evolution of control methodologies to meet the demands of intricate aerial systems.

The review delved into stability considerations, emphasizing the pivotal role of PID controllers in maintaining stability while acknowledging the challenges posed by nonlinearities, extreme operating conditions, parameter variability, and actuator dynamics. The discussion concluded by recognizing ongoing research efforts focused on integrating adaptive control, robust strategies, and machine learning to enhance stability and broaden the applicability of PID controllers in the ever-changing landscape of quadrotor dynamics.

As we navigate the complexities and embrace the opportunities presented by the intersection of UAVs, PID controllers, and advanced control strategies, this review serves as a compass, guiding researchers, engineers, and enthusiasts toward a deeper understanding and continual improvement in the field of quadrotor control. The journey continues, marked by the pursuit of innovative solutions and the quest for optimal control methodologies in the fascinating realm of unmanned aerial systems.